\documentclass[runningheads]{llncs}
\usepackage{graphicx}
\usepackage{amsmath,amssymb} 
\usepackage{color}
\usepackage[width=122mm,left=12mm,paperwidth=146mm,height=193mm,top=12mm,paperheight=217mm]{geometry}
\begin{document}
\pagestyle{headings}
\mainmatter

\title{Deep Motion Boundary Detection} 

\titlerunning{Yin et al.}

\authorrunning{Yin et al.}

\author{Xiaoqing Yin\textsuperscript{1,2}, Xiyang Dai\textsuperscript{3}, Xinchao Wang\textsuperscript{4}, \\ Maojun Zhang\textsuperscript{2}, Dacheng Tao\textsuperscript{1}, Larry Davis\textsuperscript{3}}

\institute{\textsuperscript{1}University of Sydney, \textsuperscript{2}National University of Defense Technology, \\ \textsuperscript{3}University of Maryland, \textsuperscript{4}Stevens Institute of Technology }


\maketitle

\begin{abstract}
Motion boundary detection is a crucial yet challenging problem.
Prior methods focus on analyzing the gradients and distributions of optical flow fields, 
or use hand-crafted features for motion boundary learning. 
In this paper, we propose the first dedicated end-to-end deep learning approach for 
motion boundary detection, which we term as MoBoNet.
We introduce a refinement network structure which takes
source input images, initial forward and backward optical flows as well as corresponding warping errors 
as inputs and produces 
high-resolution motion boundaries.
Furthermore, we show that the obtained motion boundaries,
through a fusion sub-network we design, 
can in turn guide the optical flows for removing the artifacts.
The proposed MoBoNet is generic and works with any 
optical flows. 
Our motion boundary detection and the refined optical flow estimation
achieve results superior to the state of the art.

\end{abstract}

\section{Introduction}
As one of the most significant cues in human visual system, 
optical flow is of great importance for visual learning, structure perception, and self-localization~\cite{brox2011large,baker2011database}. Estimating precise pixel-wise motion from video sequences in the form of optical flow is a fundamental pre-processing step for varieties of other tasks, including object detection \cite{zhou2005modified}, video super-resolution~\cite{mitzel2009video} and video denoising~\cite{buades2016patch}. 
Optical flow estimation produces a field of vectors indicating the motion of pixels between two frames,
which usually results in large segments of smooth regions.


Motion boundaries are defined as the discontinuities of such smooth regions~\cite{weinzaepfel2015learning}. 
In other words,  they correspond to the areas with sharp changes in the optical flow field
and thus divide the flow segments.
They provide vital clues for segmenting moving objects and  
many other higher-level computer vision tasks, like action recognition \cite{wang2013dense}, motion segmentation  \cite{tekalp2015digital}, anomaly detection \cite{basharat2008learning}, and object tracking \cite{yilmaz2006object}. Robust motion boundary estimation is therefore an imperative task.

Earlier motion boundary detection methods focus on gradient analysis of optical flow field~\cite{wang2013dense,papazoglou2013fast}. 
The major drawbacks of these methods are their strong sensitivity 
to the accuracy of preliminary edge detection, 
as well as their inapplicability to real-world textured videos.  
The recent work of~\cite{weinzaepfel2015learning}, Learning to Detect Motion Boundaries~(LDMB),
has proved that learning based approach 
can better explore the intrinsic relationship between multiple local features and motion boundaries,
and achieves state-of-the-art results on large-scale datasets.
However, it heavily relies on manually-designed features,
which may have limited expressive power and generalization capability.

\begin{figure}[t]
	\centering
	\includegraphics[width=1\textwidth]{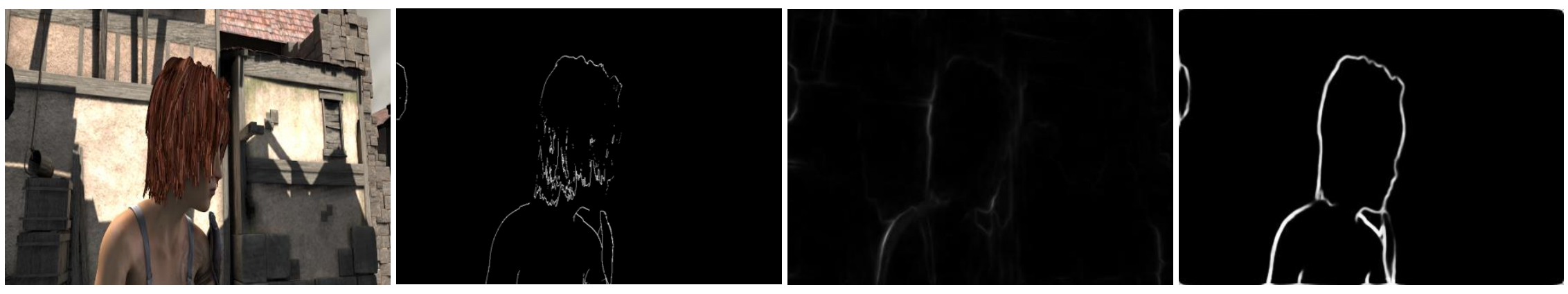}
	\caption{From left to right: input source image, motion boundary ground-truth, motion boundary prediction by LDMB \cite{weinzaepfel2015learning} and by the proposed method.}
	\label{fig1}
\end{figure}

In this paper, we propose the first dedicated deep approach for automatic motion boundary detection,
termed Motion Boundary Network~(MoBoNet), that is end-to-end trainable.
MoBoNet takes as input source images, flow estimations and warping errors,
and directly produces the corresponding motion boundaries.
We design MoBoNet as a cascade network with
residual connections, where the gradients
may propagate fast and end-to-end trainings can be
performed efficiently for  all component networks.
MoBoNet explores features of  multiple levels
and learns to analyze scenes with different magnitudes and directions of motions,
and is resistant to failures of the initial optical flow estimation. 

We show in Fig.~1 comparative visualization results of MoBoNet 
and state-of-the-art LDMB~\cite{weinzaepfel2015learning}.
As can be seen, our method  
produces more complete, more consistent and shaper
boundaries with less noise as compared to LDMB.
Also, MoBoNet is resistant to the errors of the input optical flow.
In Fig.~4 and Fig.~5, for example, our method can generate high-quality motion boundaries despite the noisy input of optical flow.

MoBoNet achieves motion boundary detection accuracies higher than current
state of the art. To further validate the utility of the obtained motion boundaries,
we reversely feed them to a sub-network we design, called fusion network, 
and test their impact on refining the optical flows.
We show that the  boundaries produced by MoBoNet indeed
benefit the optical flow estimation by generating shaper contours
and removing artifacts, leading to state-of-the-art  flow estimation results.

Our contribution is therefore the first dedicated end-to-end deep network for motion boundary detection.
The proposed approach, MoBoNet, achieves results superior to the state of the art.
We also show that, the obtained  motion boundaries can be readily applied to in turn guide and refine the
initial optical flows that are used to compute the boundaries.
This is achieved by reversely feeding the motion boundaries as input to a fusion sub-network we design.
The refined optical flows also achieve state-of-the-art performance.

\section{Related Work}
Deep methods have recently been widely applied for optical flow estimation and have achieved promising performance on challenging benchmarks. 
However, to our best knowledge, there is no existing deep-learning based models for motion boundary estimation,
which is vital  for many video analytics tasks like moving object segmentation
and optical flow enhancement. In what follows, we briefly review methods for
motion boundary estimation and related tasks, including video motion segmentation,
occlusion boundary detection, and edge detection.

\subsubsection{Motion Boundary Estimation}
Spoerri~et~al.~\cite{spoerri1990early} developed a motion boundary detection 
method for segmenting motion boundaries in synthetic footage, which analyzes bimodal distributions of 
local flow histograms and performs a structural saliency based post-processing step.  
Fleet~et~al.~\cite{fleet2000design} attempted to fit a local linear parameterized 
model for motion analysis, and to detect motion boundaries based on the observation 
that larger fitting errors correspond to motion discontinuities. 
Black and Fleet~\cite{black2000probabilistic} further enhanced this work using a probabilistic framework, where translational motion and motion discontinuity are employed to model local image patches. 
Liu~et~al.~\cite{liu2007analysis} proposed to detect motion boundaries by tracking and grouping hypothetical motion edge fragments in a bottom-up manner. 
These approaches rely on simple models and manual feature selection, 
and often fail to characterize the complicated intrinsic nonlinear correlations 
between local image features and motion boundaries.

Philippe Weinzaepfel~et~al.~\cite{weinzaepfel2015learning} proposed an effective 
learning-based solution for motion boundary detection, using 
ground truth provided by the MPI-sintel dataset \cite{butler2012naturalistic}.
In their work, both static appearance features and temporal features, including colour, optical flow, image warping errors, as well as backward flow and error, contribute to the final motion boundary results. A patch-level feature representation is built on the concatenation of all these features, combining the cues of appearance, motion and confidence in motion. This feature representation is further processed by a structured random forest model, where each tree in this model takes a patch as input and generates a boundary patch. However, the performance of motion boundary detection depends on the manually designed features. 
Furthermore, each local patch is processed independently without any interaction. 
Therefore, it still remains a challenge for this approach 
to effectively exploit varieties of motion and appearance features, 
and to further reduce the errors of motion boundary detection. 
Li~et~al.~\cite{li2016unsupervised} proposed an unsupervised learning approach for edge detection. 
This method utilizes two types of information as input: motion information in the form of noisy semi-dense matches between frames, 
and image gradients as the knowledge for edges. In the training process, motion estimation and edge detection improvement are alternated in turn. However, the performance of motion boundary estimation is limited by several issues
like the removal of weak image edges as well as label noises.

\subsubsection{Video Motion Segmentation}
Approaches for video motion segmentation 
generate optical flows and motion layers in a joint estimation framework~\cite{brox2006variational,sun2013fully}. 
However, these methods have to solve complex minimization of non-convex energy functions, leading to 
unreliable estimation results for common yet challenging scenarios, like videos with fast motions,
large displacements and compression artifacts. 
Also, the motion layer segmentation problem by itself can be ill-defined, 
as there are cases where motion boundaries form non-closed regions.


\subsubsection{Occlusion Detection and Boundaries} 
The related task of occlusion boundary detection has recently received wide attentions \cite{hoiem2011recovering,humayun2011learning,sundberg2011occlusion}. Occlusion boundaries refer to depth discontinuities. As occlusion boundaries can cause differences in optical flow (for example, when the camera or the objects move), they are equivalent to motion boundaries in some cases.
Most occlusion boundary detection methods are based on image  oversegmentation followed by a merging procedure \cite{hoiem2011recovering,sundberg2011occlusion}. In some recent works, occlusion detection tasks is combined with edge detection and a learning framework a designed to solve this problem \cite{humayun2011learning}. 

\subsubsection{Edge Detection} 
Recent edge detection methods rely on deep architectures to extract edges, 
contours and boundary features from the source images and 
achieve superior performance over conventional methods~\cite{xie2015holistically,yang2016object}.
This also inspires us to design our deep motion boundary detection network. 
Instead of exploring only the image features in edge detection task, we build our network based on varieties of inputs including initial optical flow and the source image, and explore the feature representation for efficient motion boundary detection.

\section{Methods}
Our proposed MoBoNet consists of two modules.
The first module, termed motion boundary refinement network~(refineNet), takes as input
raw input images, initial forward and backward flow estimations and warping errors and produces motion 
boundary detections. It consists of multiple sub-networks to integrate information 
from different resolutions via multiple long-range connections.
The second module, termed flow-boundary fusion network, in turn
takes motion boundaries and initial forward optical flows as input and produces 
enhanced optical flows.
Note that MoBoNet can take any optical flows estimations as input.

We show in Fig. 2 the architecture of MoBoNet.
In what follows, we describe the motion boundary refineNet in Sec. 4.1
and the flow-boundary fusion network in Sec. 4.2.

\begin{figure}[t]
	\centering
	\includegraphics[width=1\textwidth]{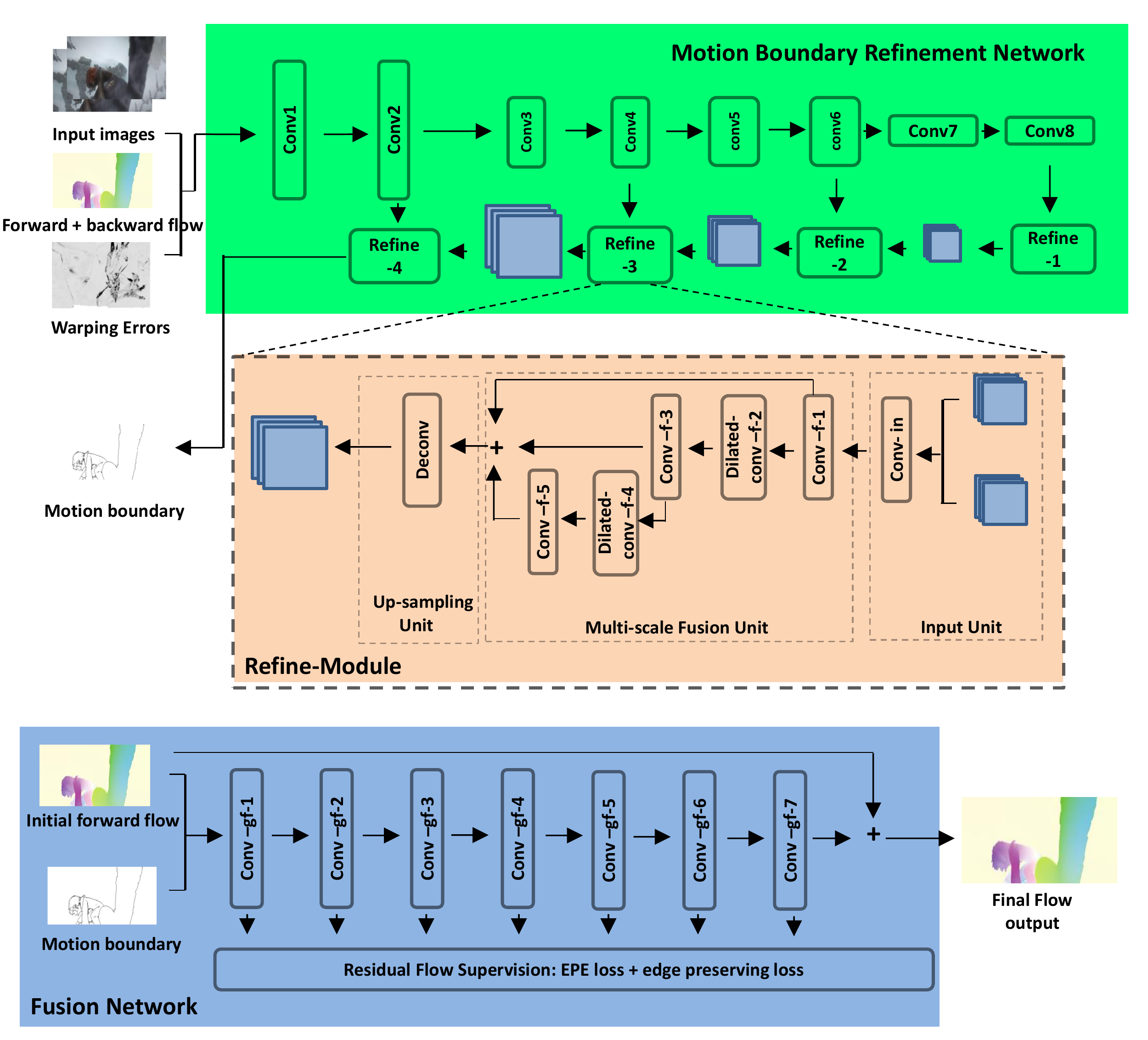}
	\caption{
		The architecture of MoBoNet. It comprises two
		modules, the motion boundary refineNet that estiamtes 
		motion boundary detections and the flow-boundary fusion network
		that reversely enhances optical flows using motion boundaries.}
	\label{fig2}
\end{figure}




\subsection{Motion Boundary RefineNet}
In this section, we provide detailed descriptions of the proposed Motion Boundary RefineNet, 
which utilizes a refinement network architecture to 
generate multi-scale feature maps for motion boundary estimation.
Three types of inputs are adopted for motion boundary prediction: 
input image containing edges and structural information, 
forward optical flows and corresponding warping errors, 
as well as backward optical flows and corresponding warping errors. 
The refineNet thus takes as input various information including 
image evidences, initial motion estimations and temporal cues


The proposed Motion Boundary RefineNet consists of two components: 
forward contraction and backward refinement. 
The forward pathway generates high-dimensional low-resolution feature maps with appearance and motion information. 
In the backward refinement pathway, these feature maps are fused with intermediate features along the forward pass. The refinement part consists of four cascaded sub-net blocks (Refine - 1 to 4 in Fig. 2). 
Each of these blocks takes the output of the previous block and feature maps from the contraction part as input, and feeds its output to the following block.
The spacial resolution of feature maps is increased by a factor 2 by each sub-net 
until the input resolution is reached eventually. 
Each sub-network in the refinement path contains three units: input unit, multi-scale fusion unit and up-sampling unit, 
which we will describe in detail as follows.



1. Input unit: 
Each input unit takes two inputs: refinement feature maps from the previous refinement module, and intermediate feature maps from the forward pass. These two inputs are concatenated and processed by a convolution layer (conv-in in Fig. 2). 

2. Multi-scale fusion unit: The resulting feature maps from the input unit are then transferred to the multi-scale fusion unit. Context information for a larger receptive range is extracted, where features are efficiently processed by increasing window sizes and fused together to generate output feature maps. Our motivation stems from the deep method that performed multi-scale feature fusion for edge learning \cite{xie2015holistically}, which is insensitive to the scales of input images.

We construct this unit as a combination of convolution layers conv-f-1 to conv-f-5, where conv-f-2 and conv-f-4 are dilated convolution layers that are used to enlarge the receptive field without using pooling. Each convolution layer is followed by a ReLU layer. Features from conv-f-1 are re-used by conv-f-2 and conv-f-3 for further processing from a larger receptive field. Similarly, features from conv-f-3 are re-used by conv-f-4 and conv-f-5 on an even larger scale. Finally, all the features maps of different scales are fused by summation. In addition, multiple paths are built between the input and output of the Multi-scale fusion unit, which is helpful for learning complex features. The residual paths within this unit further boost gradient propagation in the training process. 

3. Up-sampling unit: After multi-scale fusion, the refinement module expands the resolution of feature maps by a deconvolution layer. The output feature maps are then used by the following refinement sub-network.

Detailed configuration of the motion boundary refineNet is shown in Tab. 1. All the input images are resized to [height, width] = [320, 448].




\begin{table}
	\renewcommand{\arraystretch}{1.3}
	\caption{Detailed conguration of the refineNet. }
	\label{table_example}
	\centering
	\begin{tabular}{|c|c|c ||c|c|c|}
		\hline
		{\bf Name} & {\bf Kernel} & {\bf Output Size} & {\bf Name} & {\bf Kernel} & {\bf Output Size}\\ 
		\hline
		conv1,2 & 3$\times$3 & 320$\times$448$\times$64  & deconv5 & 2$\times$2 & 40$\times$56$\times$256\\
		\hline
		pool1     & 2$\times$2 & 160$\times$224$\times$64 & refine-1 convs & 3$\times$3 & 40$\times$56$\times$256\\
		\hline
		conv3,4 & 3$\times$3 & 160$\times$224$\times$128 & deconv4 & 2$\times$2 & 80$\times$112$\times$128\\
		\hline
		pool2     & 2$\times$2 & 80$\times$112$\times$128 & refine-2 convs & 3$\times$3 & 80$\times$112$\times$128\\ 
		\hline
		conv5,6, 1-2 & 3$\times$3 & 80$\times$112$\times$256 & deconv3 & 2$\times$2 & 160$\times$224$\times$64\\ 
		\hline
		pool3     & 2$\times$2 & 40$\times$56$\times$256 & refine-3 convs & 3$\times$3 & 160$\times$224$\times$64\\ 
		\hline
		conv7,8, 1-2 & 3$\times$3 & 40$\times$56$\times$512 & deconv2 & 2$\times$2 & 320$\times$448$\times$32\\ 
		\hline
		pool4     & 2$\times$2 & 20$\times$28$\times$512 & refine-4 convs & 3$\times$3 & 320$\times$448$\times$32\\ 
		\hline
	\end{tabular}
\end{table}


During the training process for the motion boundary refineNet, we aim to minimize the class-balanced cross entropy loss $L_c$ \cite{xie2015holistically} between the output motion boundary and the ground truth motion boundary: 
\begin{equation}
L_c=-\beta\sum_{j\in Y_+}\log{Pr(y_j=1|X;W)}-(1-\beta)\sum_{j\in Y_-}\log{Pr(y_j=0|X;W)}
\end{equation}
where $\beta = | Y_-|/|Y|$ and $1-\beta = |Y_+|/|Y|$. $|Y_-|$ and $|Y_+|$ denote the boundary and non-boundary ground truth label sets, respectively.

\subsection{Flow-Boundary Fusion Network}

Our second module, flow-boundary fusion network or simply fusion network,
fuses the motion boundaries and initial optical flows and produces strengthened 
flow results. Our motivation stems from the fact that motion boundaries 
can provide valuable cues for optical flow estimation and 
add extra prior knowledge to regularize the optical flow 
optimization process.


The fusion network also follows
the residual network design as shown in Fig.~2.
Since the initial optical flow contains most of the necessary details, 
the edge-guided residual network only need to focus on estimating the 
difference between the filtered flow and the initial flow, 
especially the high-frequency regions along the motion boundaries. 
Specifically, we write this residual learning as:
\begin{equation}
f_1 = F(f_0, m, {W_i}) + f_0
\end{equation}
where $f_1$ and $f_0$ are the final output flow and the initial flow respectively,
$W_i$ corresponds to the network parameters and $m$ demotes the input motion boundaries. 
Our goal here is to only predict a residual flow $F(f_0, m, {W_i})$ between $f_1$ and $f_0$,
as most values in the residue are likely to be zero or small.


The proposed fusion network consists of 8 convolutional layers, each applying 5 $\times$ 5 kernel and 64 feature maps followed by ReLU as the activation function. Zero-padding is adopted to maintain the dimension at each layer.
We concatenate output feature maps of layer $k$ with the $k$-th flow residue prediction and use this concatenation as the input for layer $k+1$.

We transform the activation of each layer into a flow residue prediction using a single convolution layer. Each loss layer takes three inputs: residual flow prediction, initial flow and the ground truth flow. Each of the flow residues are added to the initial flow and then compared with the ground truth flow. 
As EPE loss~\cite{brox2011large,weinzaepfel2013deepflow,dosovitskiy2015flownet} 
used in previous optical flow estimation methods does not necessarily guarantee high accuracy, 
we further utilize the gradient of optical flow obtained from Prewitt filter, 
and enforce the network to preserve high frequency structures in the flow results. 
We implement this boundary preserving loss by defining the Prewitt filter as a convolutional 
layer with fixed filters. 
We write the final loss function as a combination of EPE and boundary preserving:
\begin{equation}
L_{epe} = \frac{1}{n}\left(\sum_{i,j}EPE(F_{i,j},F_{i,j}^{gt})+\sum_{i,j}\|P(F_{i,j})-P(F_{i,j}^{gt})\|_2^2\right)
\end{equation}
where $P$ denote the Prewitt filter, $F$ and $F^{gt}$ represent the estimated flow and the corresponding ground truth, and $n$ is the number of pixels.

\subsection{Training Process}
We implement our model in Caffe \cite{jia2014caffe} and apply the adaptive gradient algorithm (ADAGRAD) \cite{duchi2011adaptive} to optimize the entire network. We first train the motion boundary refineNet to learn the mapping relationship between the inputs and motion boundary results. Network parameters are learned through minimization of the class balanced cross entropy loss function between the estimated motion boundary and the corresponding ground truth. We set the initial learning rate as 1e-4 and reduce it by a factor of 2 every 100K iterations. The motion boundary detection network is first trained on the Flythings3D dataset~\cite{mayer2016large} with total iterations of 500K. To test our model on the sintel~\cite{butler2012naturalistic} and kitti dataset~\cite{geiger2012we} , we further finetune our network with 60K iterations on the training sets provided by the two datasets, respectively. We randomly select half of the sequences for training, and test on the other half, as done by LDMB \cite{weinzaepfel2015learning}. And we use the model finetuned on sintel dataset for testing on Middleburry and YMB dataset.

Next, we fix the paramters in the motion boundary detection network, and perform training for the flow-boundary fusion network. The initial learning rate is set as 1e-3 and reduced by a factor of 10 every 50K iterations. During this training step, the learned motion boundaries propagate prior information to the final flow result. Similar to the motion boundary detection network, this fusion network is first trained on the Flythings3D dataset with 300K iterations, and then finetuned on the sintel and kitti datasets with 50K iterations.

In our experiments, varieties of data augmentations are performed, including translation, rotation, scaling, additive Gaussian noise and changes on brightness, contrast, gamma and color. Detailed augmentation strategies are given as follows: translation [-15\%, 15\%] for image width in x and y, rotation [$-20^{\circ}$, $20^{\circ}$], scaling [0.8, 1.8], Gaussian noise with sigma uniformly sampled [0, 0.05], multiplicative color changes [0.8, 1.6], gamma values [0.6, 1.6] and additive brightness changes (Gaussian) with a sigma value of 0.15.

\section{Experiments}
In this section, we discuss our experimental setup and results. We first introduce the datasets and evaluation protocols in Section 4.1 and then compare our motion boundary detection results with state-of-the-art methods qualitatively and quantitatively in Section 4.2. We further show the results of optical flow optimization in Section 4.3, which demonstrates the effectiveness of guidance from motion boundary.  

We apply three widely used optical flow estimation algorithms for initial flow generation: FlowNet \cite{dosovitskiy2015flownet}, DC-Flow \cite{xu2017accurate} and Deep-Flow \cite{weinzaepfel2013deepflow}.
For each of these flows, we train a separate model. 


\subsection{Datasets and Evaluation Protocols}

\subsubsection{Motion Boundary datasets} 
We train our model on Flythings3D dataset, and perform finetuning on Sintel and KITTI dataset. Our framework is evaluated on motion boundary datasets generated from the optical flow datasets, and YMB dataset. 

\textbf{Flythings3D}: In Flythings3D dataset \cite{mayer2016large}, multiple types of simulated objects are automatically generated and combined to form 25 000 stereo frames, with each object randomly scaled, rotated, textured and then placed along randomized 3D trajectory. Both ground truth optical flow and motion boundaries are provided, which is essential for training the proposed motion boundary detection network.

\textbf{YMB}: YouTube Motion Boundaries dataset (YMB) \cite{weinzaepfel2015learning} contains 60 sequences captured from real-world scenes including variaties of persons, objects and poses. In each sequence, motion boundaries in one frame are provided by three independent annotators.

We also evaluate motion boundary detection on general optical flow datasets: Sintel, KITTI and Middleburry.
Binary motion boundary ground truth is computed from corresponding optical flow datasets to further evaluate the performance of our motion boundary detection method. We follow the strategy of \cite{weinzaepfel2015learning}, which applies various thresholds of flow gradient for generating the boundaries.

\subsubsection{Optical Flow datasets}
To demonstrate the guidance of motion boundary on optical flow refinement, we further evaluate our framework on optical flow benchmarks: Sintel, KITTI and Middlebury.

\textbf{Sintel}: MPI-Sintel dataset \cite{butler2012naturalistic}  is generated from an animated movie which contains multiple sequences with various motions. Two versions of this dataset, ‘Clean’ and ‘Final’, are provided, with the latter contains more realistic image effects including motion blur and atmospheric effects. In our experiments, we exploit both the Clean and Final versions to train and test our model.

\textbf{KITTI}: The KITTI 2012 dataset \cite{geiger2012we} contains 194 real-world image pairs. All the scenes are simultaneously recorded by a camera and a 3D laser scanner to obtain the ground truth optical flow. Thus the scenes are assumed to be rigid and the motions stem from a moving observer.

\textbf{Middlebury}: The Middlebury dataset is composed of 8 training sequences, each of which contains 2 to 8 frames and ground-truth optical flow provided for the central frame. Most motions in the sequences are limited to small displacements, making the average endpoint error for recent optical flow estimation methods lower than the former two datasets.

\subsubsection{Evaluation Protocol}
We apply the precision-recall criterion to evaluate the results of motion boundary detection, with the evaluation code provided by edge detection benchmark BSDS \cite{martin2001database}. Pixel-wise recall and precision curves are calculated using both ground-truth and prediction results of motion boundary prediction. 

\subsection{Motion Boundary Detection}

\subsubsection{Quantitative Evaluation}

We run the proposed model as well as LDMB \cite{weinzaepfel2015learning} for each of the initial flow estimation method, and  
report mean Average-Precision (mAP) for all datasets in
Tab. 2. 

Compared with baseline motion boundary detection methods, the proposed motion boundary net achieves superior performance on all the motion boundary datasets. We use the model finetuned on MPI-sintel to compute the results on Middlebury and YMB datasets. Although the proposed network is trained on synthetic datasets, it generalizes well to other real -world datasets and performs well without specific tuning. 

In contrast to the previous method that rely on hand-crafted features \cite{weinzaepfel2015learning} or simple analysis on initial optical flow, the proposed method performs robust motion boundary detection and achieve more accurate results. DC-Flow \cite{xu2017accurate} gives the best performance on all datasets. This can be explained by the sharpness of the flow boundaries thanks to the high quality-optical flow estimation result. 
As Middlebury dataset mainly contains small motions that can be easily estimated , the initial flow estimations are more close to the ground truth. Therefore, motion boundaries are predicted with higher accuracy.

\begin{table}
	\renewcommand{\arraystretch}{1.3}
	\newcommand{\tabincell}[2]{\begin{tabular}{@{}#1@{}}#2\end{tabular}}
	\caption{Comparison of the performance (mAP) of our approach for different input flows.}
	\label{table_example}
	\centering
	\begin{tabular}{|c|c|c|c|c|c|}
		\hline
		& Middleburry& MPI-sintel clean & MPI-sintel final &KITTI & YMB \\
		\hline
		FlowNet\cite{dosovitskiy2015flownet}+LDMB\cite{weinzaepfel2015learning} & 82.5 & 68.4 &  59.7 & 62.6 & 64.3 \\
		\hline
		FlowNet\cite{dosovitskiy2015flownet}+Propoesd& \bf{87.2} & \bf{75.7} & \bf{67.9} & \bf{68.7} & \bf{70.6} \\
		\hline
		DeepFlow\cite{weinzaepfel2013deepflow}+LDMB\cite{weinzaepfel2015learning} & 89.0 & 75.8 & 67.7 & 65.2& 68.6 \\
		\hline
		DeepFlow\cite{weinzaepfel2013deepflow}+Proposed & \bf{92.7} & \bf{79.1} & \bf{71.8} & \bf{69.6} &\bf{72.9} \\
		\hline
		DC-Flow\cite{xu2017accurate}+LDMB\cite{weinzaepfel2015learning} & 94.2 & 83.2 & 75.6 & 74.3 & 77.5 \\
		\hline
		DC-Flow\cite{xu2017accurate}+Proposed & \bf{95.6} & \bf{85.4} & \bf{78.1} & \bf{75.7} & \bf{79.1} \\
		\hline
	\end{tabular}
\end{table}

Fig. 3 shows the precision recall curves for different datasets using FlowNet \cite{dosovitskiy2015flownet} to generate initial flow. Each point of the curves corresponds to a different threshold on the strength of predicted motion
boundary. For example, higher threshold can lead to fewer predicted pixels, i.e., lower recall and higher precision. In order to avoid problems related to over/underassignment of ground-truth and predicted pixels, a standard non-maximal suppression technique is applied to obtain thinned edges for evaluation, as done in LDMB \cite{weinzaepfel2015learning}. The proposed framework achieves the best result compared to previous methods. Although performance is reported for FlowNet \cite{dosovitskiy2015flownet}, all the other flow estimation methods behave in a similar manner. 

\begin{figure}[htb]  
	\centering  
    \includegraphics [width=1\textwidth,height=0.6\textwidth]{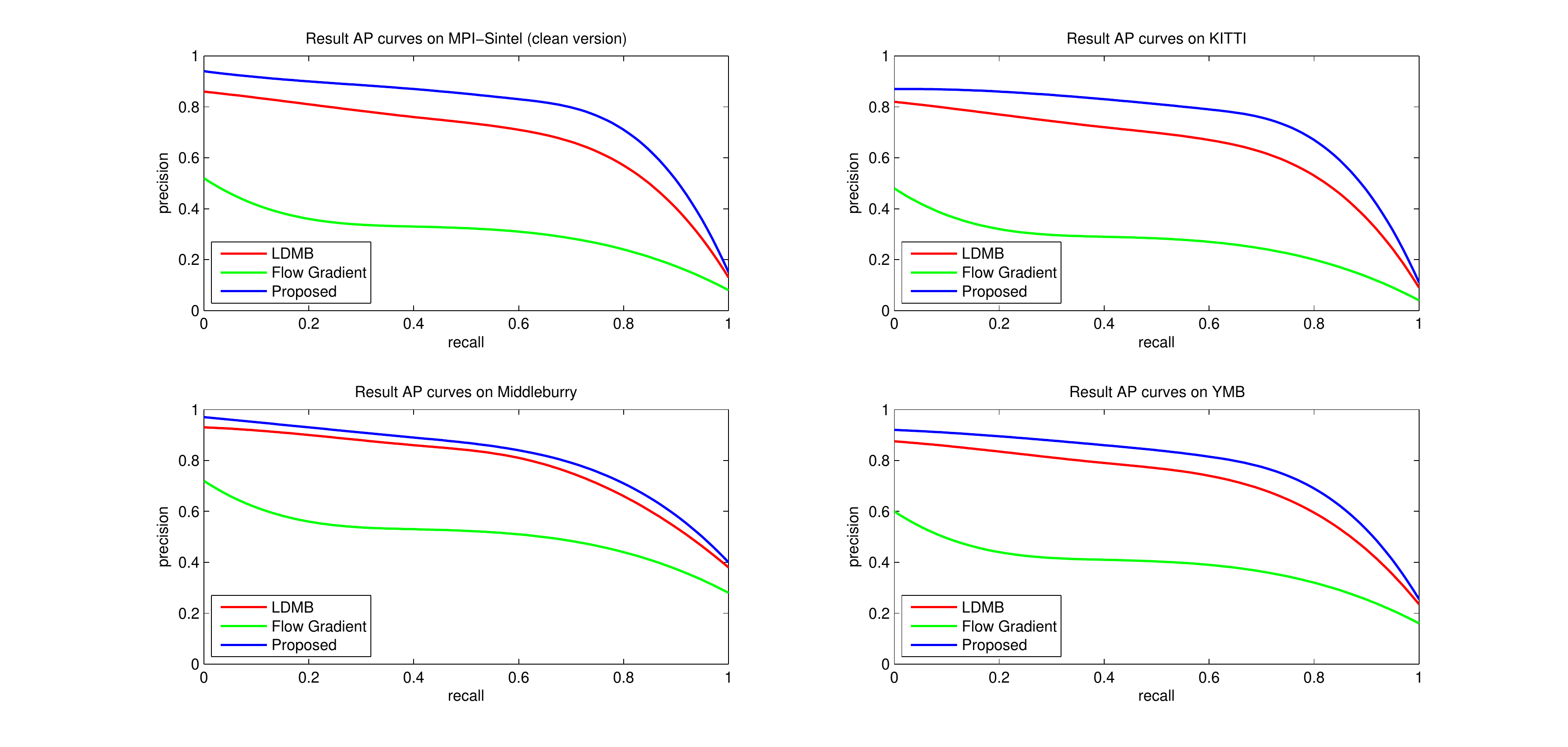}
	\caption{Precision-Recall curves on different datasets for FlowNet \cite{dosovitskiy2015flownet}. Our proposed method achieves the best result compared to previous methods. Although performance is reported for FlowNet \cite{dosovitskiy2015flownet}, all the other flow estimation methods behave in a similar manner.} 
	\label{1}  
\end{figure}

\subsubsection{Qualitative Evaluation}

Fig. 4 provides visual comparisons for norm of flow gradient, LDMB \cite{weinzaepfel2015learning} and our predictions. Our results show the best visual quality. While results obtained by flow gradient and LDMB\cite{weinzaepfel2015learning} contain significant noise, our method shows a clear result with better consistency and less errors. Some background edges are identified as motion boundaries in LDMB\cite{weinzaepfel2015learning}, while this problem is avoided in our results.


Furthermore, Fig. 4 shows the robustness of the proposed method to inaccurate or oversmooth optical flow inputs. Noise is observed in the initial forward flow in the first column, and the flow of the character oversmooth to the background in the fourth column. In spite of the flow estimation errors and the spreading of motions, motion boundaries can be precisely estimated by the proposed method. Multiple types of features are fed to the proposed motion boundary detection network, which helps to provide both appearance and flow information for motion boundary detection and overcome the errors of initial optical flow estimation.

\begin{figure}[htbp]
	\centering
	\includegraphics [width=1\textwidth]{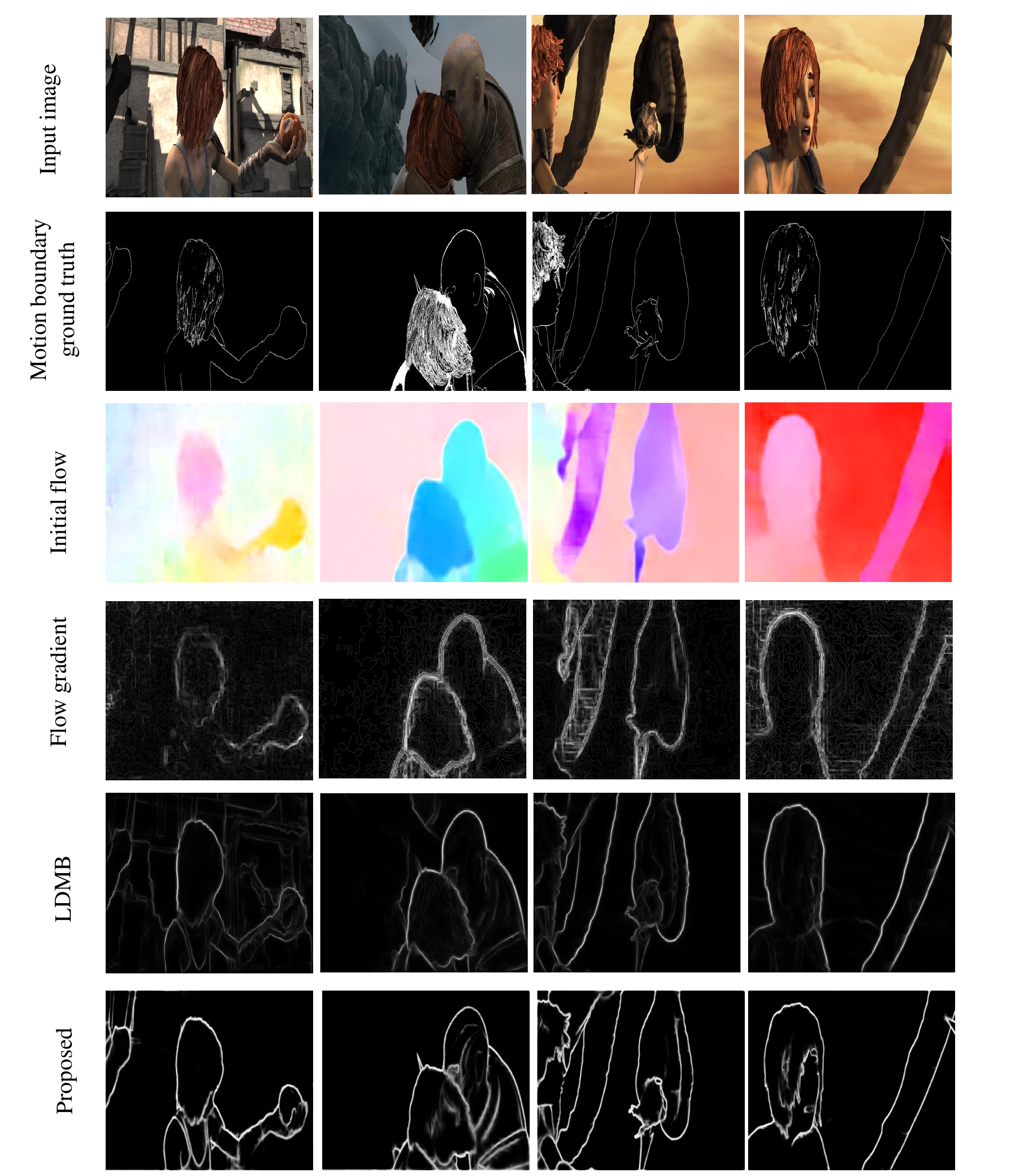}
	\caption{Qualitative results on our synthesized datasets. From top to down, we show the input image,  motion boundary ground truth, initial forward optical flow, results of norm of flow gradient, results of state-of-the-art method\cite{weinzaepfel2015learning}, and the results of the proposed approach. Our method achieves the best overall visual quality. }
\end{figure}

\subsection{Optical Flow Optimization}

\subsubsection{Quantitative Evaluation}

The optimization results for optical flow are listed in Tab. 3. With the guidance of motion boundaries, improvements on optical flow estimation are observed on all the tested methods.


\begin{table}
	\renewcommand{\arraystretch}{1.3}
	\newcommand{\tabincell}[2]{\begin{tabular}{@{}#1@{}}#2\end{tabular}}
	\caption{Average endpoint errors (in pixels) compared on different datasets.}
	\centering
	\begin{tabular}{|c|c|c|c|c|}
		\hline
		& MPI-sintel clean & MPI-sintel final & KITTI 2012 & Middleburry  \\
		\hline
		DeepFlow\cite{weinzaepfel2013deepflow} & 2.66 & 3.57 & 4.48 & 0.25  \\
		\hline
		DeepFlow\cite{weinzaepfel2013deepflow}+Proposed & 2.39 & 3.14 & 4.11 & 0.22  \\
		\hline
		DC-Flow\cite{xu2017accurate} & 1.84 & 2.76 & 4.94 & 0.32 \\
		\hline
		DC-Flow\cite{xu2017accurate}+Proposed & 1.68 & 2.59 & 4.77 & 0.31 \\
		\hline
		FlowNet\cite{dosovitskiy2015flownet} & 4.31 & 5.87 & 9.35 & 1.15 \\
		\hline
		FlowNet\cite{dosovitskiy2015flownet}+Proposed & 3.86 & 4.96 & 8.04 & 0.98 \\
		\hline
	\end{tabular}
\end{table}

\subsubsection{Qualitative Evaluation}
Visual comparisons for FlowNet\cite{dosovitskiy2015flownet} on MPI-Sintel clean dataset are shown in Fig. 5. The outputs of deep motion boundary detection network can provide detailed structures for optical flow filtering. By combining the motion boundaries to the initial optical flows, sharp high-frequency details are recovered while the errors in the initial flow are further reduced. In the bottom three rows in Fig.5, We show the outputs of 3rd, 6th and the final layer in the flow-boundary fusion network, respectively. Noise and artifacts in the non-boundary regions are gradually removed, while more structures are added to the final result. Note also that a few details may be removed because of the errors in motion boundary detection. For example, in the second column of Fig. 5, the small region of motion (in red) is removed due to the failure of motion boundary detection. 



\begin{figure}[htbp]
	\setlength{\abovecaptionskip}{0pt}
	\centering
	\graphicspath{figure4.pdf/}
	\includegraphics [width=1\textwidth]{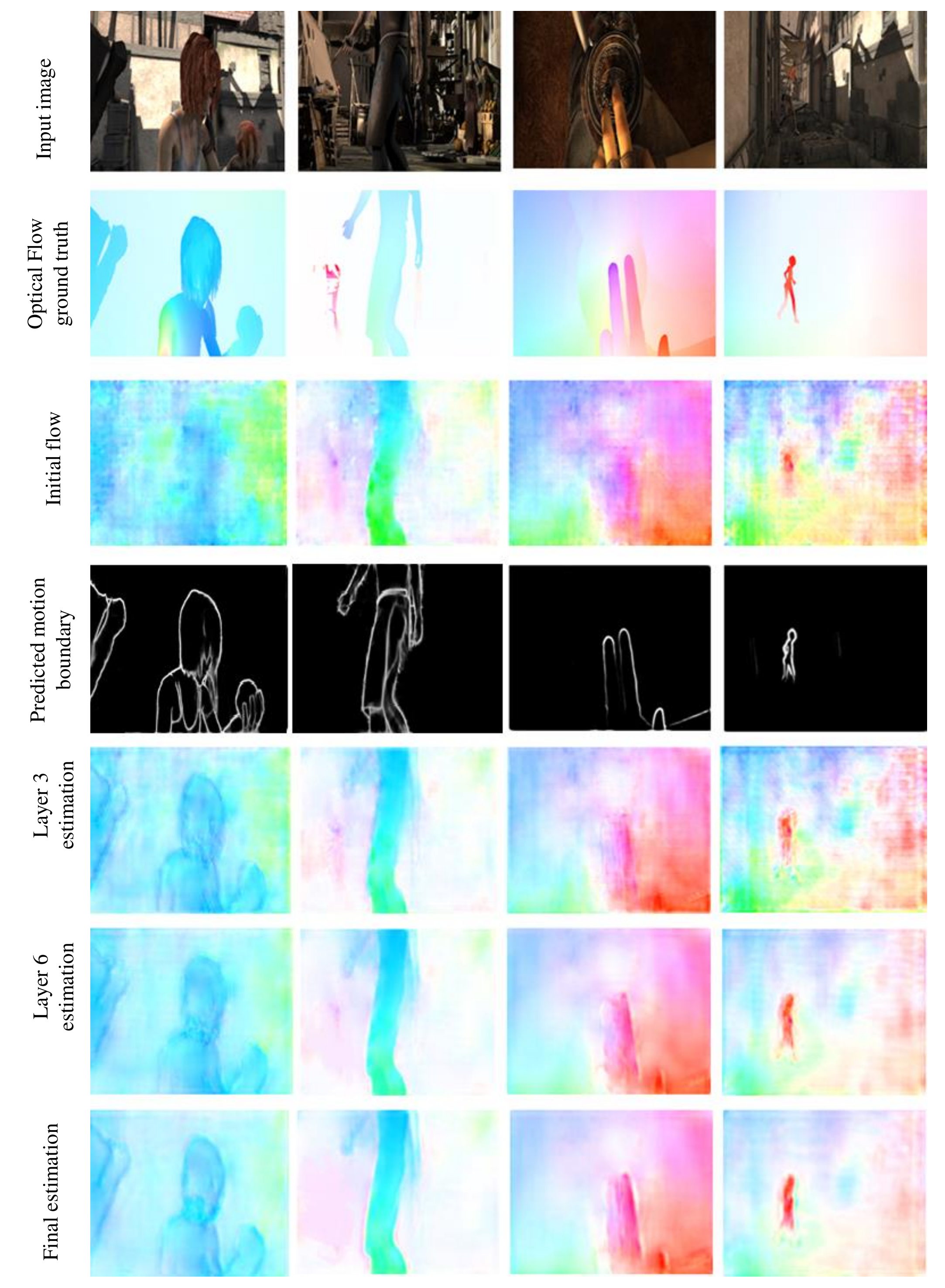}
	\caption{Qualitative results for optical flow optimization on MPI-Sintel Clean dataset. From top to down: the input images, optical flow ground truth, the initial forward flow estimation, detected motion boundaries, and the optimized results from 3rd, 6th and final layer in the flow-boundary fusion network.}
\end{figure}

\section{Conclusions}
In this paper, we proposed a joint deep learning framework for motion boundary detection and optical flow optimization. Inspired by conventional motion boundary detection methods and recent development of deep learning architectures in computer vision tasks, we propose a cascaded refinement network architecture, motion-boundary-RefineNet, to detect motion boundaries in an end-to-end manner. The proposed network explores multiple levels of features from both the initial optical flow and source image, and achieves effective training with residual connections employed in all component networks. 
Based on the motion boundary results, a fusion network is proposed, which  exploits the guidance of motion boundary to perform boundary-respecting filtering for the initial flow and further remove artifacts. 
The effectiveness of the proposed framework is demonstrated by extensive experiments on motion boundary detection and optical flow estimation, both qualitatively and quantitatively.  We will further extend this framework to further improve the robustness to different input optical flow results, and apply our method to other areas including video segmentation and action recognition.

\bibliographystyle{unsrt}
\bibliography{egbib_arxiv}
\end{document}